\def\year{2021}\relax
\documentclass[letterpaper]{article} 
\usepackage{aaai21}  
\usepackage{times}  
\usepackage{helvet} 
\usepackage{courier}  
\usepackage[hyphens]{url}  
\usepackage{graphicx} 
\usepackage{pgfplots}
\usepackage{amsmath}
\usepackage{amssymb}
\usepackage{caption}
\usepackage{subcaption}

\definecolor{_orange}{RGB}{255, 230, 204}
\definecolor{_yellow}{RGB}{255, 242, 204}
\definecolor{_blue}{RGB}{218, 232, 252}
\definecolor{_red}{RGB}{248, 206, 204}
\definecolor{_green}{RGB}{213, 232, 212}
\definecolor{_grey}{RGB}{245,245,245}
\newcommand{\ffrac}[2]{\ensuremath{\frac{\displaystyle #1}{\displaystyle #2}}}
\newcommand{\argmax}[1]{\underset{#1}{\operatorname{arg}\,\operatorname{max}}\;}
\newcommand{\rulesep}{\unskip\ \vrule\ }
\usepackage{natbib}  
\usepackage{caption} 
\title{Unsupervised Learning of Discourse Structures using a Tree Autoencoder}
\author{
    Patrick Huber, Giuseppe Carenini\\
}
\affiliations{
    Department of Computer Science \\
    University of British Columbia \\
    Vancouver, BC, Canada, V6T 1Z4 \\
    {\tt \{huberpat, carenini\}@cs.ubc.ca}}

\begin{document}

\maketitle

\begin{abstract}
Discourse information, as postulated by popular discourse theories, such as RST and PDTB, has been shown to improve an increasing number of downstream NLP tasks, showing positive effects and synergies of discourse with important real-world applications. While methods for incorporating discourse become more and more sophisticated, the growing need for robust and general discourse structures has not been sufficiently met by current discourse parsers, usually trained on small scale datasets in a strictly limited number of domains. This makes the prediction for arbitrary tasks noisy and unreliable. The overall resulting lack of high-quality, high-quantity discourse trees poses a severe limitation to further progress. 
In order the alleviate this shortcoming, we propose a new strategy to generate tree structures in a task-agnostic, unsupervised fashion by extending a latent tree induction framework with an auto-encoding objective. The proposed approach can be applied to any tree-structured objective, such as syntactic parsing, discourse parsing and others. However, due to the especially difficult annotation process to generate discourse trees, we initially develop a method to generate larger and more diverse discourse treebanks. In this paper we are inferring general tree structures of natural text in multiple domains, showing promising results on a diverse set of tasks. 
\end{abstract}

\section{Introduction}
Discourse Parsing is a key Natural Language Processing (NLP) task for processing multi-sentential text. Most research in the area focuses on one of the two main discourse theories -- RST \cite{mann1988rhetorical} or PDTB \cite{prasadpenn}. The latter thereby postulates shallow discourse structures, combining adjacent sentences and mainly focuses on explicit and implicit discourse connectives. The RST discourse theory, on the other hand, proposes discourse trees over complete documents in a constituency-style manner, with tree leaves as so called Elementary Discourse Units (or EDUs), representing span-like sentence fragments. Internal tree-nodes encode discourse relations between sub-trees as a tuple of \{Nuclearity, Relation\}, where the nuclearity defines the sub-tree salience in the local context, and the relation further specifies the type of relationship between the binary child nodes (e.g. Elaboration)\footnote{We only generate plain discourse structures in this work, not considering nuclearity and relation labels.}. While both discourse theories are of great value to the field of NLP, and have stimulated much progress in discourse parsing, there are major drawbacks when data is annotated according to these theories: \\
\textbf{(1)} Since both theories rely on annotation-guidelines rather than data-driven algorithms, the human factor plays a substantial role in generating treebanks, posing a difficult task on linguistic experts.
In this work, we are eliminating the human component from the annotation process by employing a data-driven approach to generate discourse trees directly from natural language, capturing commonly occurring phenomena in an unsupervised manner.\\
\textbf{(2)} The annotation process following human-generated guidelines, especially following the RST discourse theory, 
is expensive and tedious, as the annotation itself requires linguistic expertise and a full understanding of the complete document. This limits available RST-style discourse corpora in both, size and number of domains where gold-standard datasets exist. Using an automated, data-driven approach as described in this paper allows us to crucially expand the size and domain-coverage of datasets annotated with RST-style discourse structures.

With the rapidly growing need for robust and general discourse structures for many downstream tasks and 
real-world applications (e.g. \citet{gerani2014abstractive, nejat2017exploring, ji2017neural, xiao-etal-2020-really, huber-carenini-2020-sentiment}), the current lack of high-quality, high-quantity discourse treebanks poses a severe shortcoming. 

Fortunately, more data-driven alternatives to infer discourse structures have been previously proposed. For example, our recently published MEGA-DT discourse treebank 
\cite{huber2020mega} with automatically inferred discourse structures and nuclearity attributes from large-scale \textit{sentiment} datasets already reached state-of-the-art (SOTA) performance on the inter-domain discourse parsing task. Similarly, \citet{liu2018learning} infer latent discourse trees from the \textit{text classification} task, and \citet{liu2019single} employ the downstream task of \textit{summarization} using a transformer model to generate discourse trees.
Outside the area of discourse parsing, syntactic trees have previously been inferred according to several strategies, e.g. \citet{socher2011semi, yogatama2016learning, choi2018learning, maillard2019jointly}.

In general, the approaches mentioned above 
have shown to capture valuable structural information. Some models outperform baselines trained on human-annotated datasets (see \citet{huber2020mega}), others have proven to enhance diverse downstream tasks \cite{liu2018learning, liu2019single, choi2018learning}. However, despite these initial successes, one critical limitation that all aforementioned models share is the task-specificity, possibly only capturing downstream-task related information. 
This potentially compromises the generality of the resulting trees, as for instance shown for the model using \textit{text classification} data \cite{liu2018learning} in \citet{ferracane2019evaluating}. 
In order to alleviate this limitation of task-specificity, we propose a new strategy to generate tree structures in a task-agnostic, unsupervised fashion by extending the latent tree induction framework proposed by \citet{choi2018learning} with an auto-encoding objective. 
Our system thereby extracts important knowledge from natural text by optimizing both the underlying tree structures and the distributed representations. We believe that the resulting discourse structures effectively aggregate related and commonly appearing patterns in the data by merging coherent text spans into intermediate sub-tree encodings, similar to the intuition presented in \citet{drozdov2019unsupervised}. However, in contrast to the approach by \citet{drozdov2019unsupervised}, our model makes discrete structural decisions, rather than joining possible subtrees using a soft attention mechanism. We believe that our discrete tree structures allow the model to more efficiently achieve the autoencoder objective in reconstructing the inputs, directly learning how written language can be aggregated in the wild (comparable to previous work in language modelling \cite{jozefowicz2016exploring}).
In general, the proposed approach can be applied to any tree-structured objective, such as syntactic parsing, discourse parsing and further problems outside of NLP, like tree-planning \cite{guo2014deep} and decision-tree generation \cite{irsoy2016autoencoder}. Yet, due to the especially difficult annotation process to generate discourse trees, we initially develop a method to 
generate much larger and more diverse discourse treebanks.

\section{Related Work}
Within the last decade, general autoencoder frameworks have been frequently used to compress data, such as in \citet{srivastava2015unsupervised}. More recently, sequential autoencoders have been applied in the area of NLP \cite{li2015hierarchical}, with many popular approaches, such as sequence-to-sequence learning models \cite{sutskever2014sequence} having strong ties to sequential autoencoders. 
Based on the promising results of the sequential autoencoder, researchers started to compress and reconstruct more general structures in tree-style models, such as \citet{chen2018tree} showing that with available gold-standard trees, the programming-language translation task (e.g. from CoffeeScript to JavaScript) can be learned with a tree-to-tree style neural autoencoder network. Furthermore, variational autoencoders have been shown effective for the difficult task of grammar induction \cite{kusner2017grammar}. 

While both previously mentioned applications for tree-style autoencoder models require readily available tree structures to guide the aggregation process, another line of work by \citet{socher2011semi} overcomes this requirement by using the reconstruction error of an autoencoder applied to every two adjacent text spans as an indicator for syntactic correctness within a sentence. In their model, \citet{socher2011semi} combine the tree-inference objective with the autoencoder topology, training an unsupervised tree-structured model, which is subsequently fine-tuned on a small-scale supervised dataset. While their model is clearly comparable to our approach, there are three major differences: (1) They make sequential, local decisions on the aggregation of spans to generate a tree structure, rather than optimizing the complete process holistically. (2) Their model uses an unsupervised objective in the initial step but requires supervision in later stages and (3) The model has been only applied to syntactic parsing. In contrast, we apply our model to discourse parsing, which arguably introduces further difficulties, as we will discuss in section \ref{discourse_adaptations}. 

Recently, \citet{choi2018learning} showed a promising approach to infer tree structures in a holistic and parallelizable manner, generating task-depended trees solely relying on sentiment-related information. In their model, they make use of the Gumbel-Softmax  \cite{jang2016categorical} (also used in similar ways in \citet{corro2018differentiable, corro2019learning}), allowing the neural network to make discrete decisions while still being able to use standard approaches like back-propagation to optimize the model. By combining a similar objective to \citet{socher2011semi} and \citet{chen2018tree}, we utilize the discrete decision-process in \citet{choi2018learning}, positioning our work at the intersection of these two lines of research.
\begin{figure*}
    \centering
    \setlength{\belowcaptionskip}{-0pt}
    \includegraphics[width=1\linewidth]{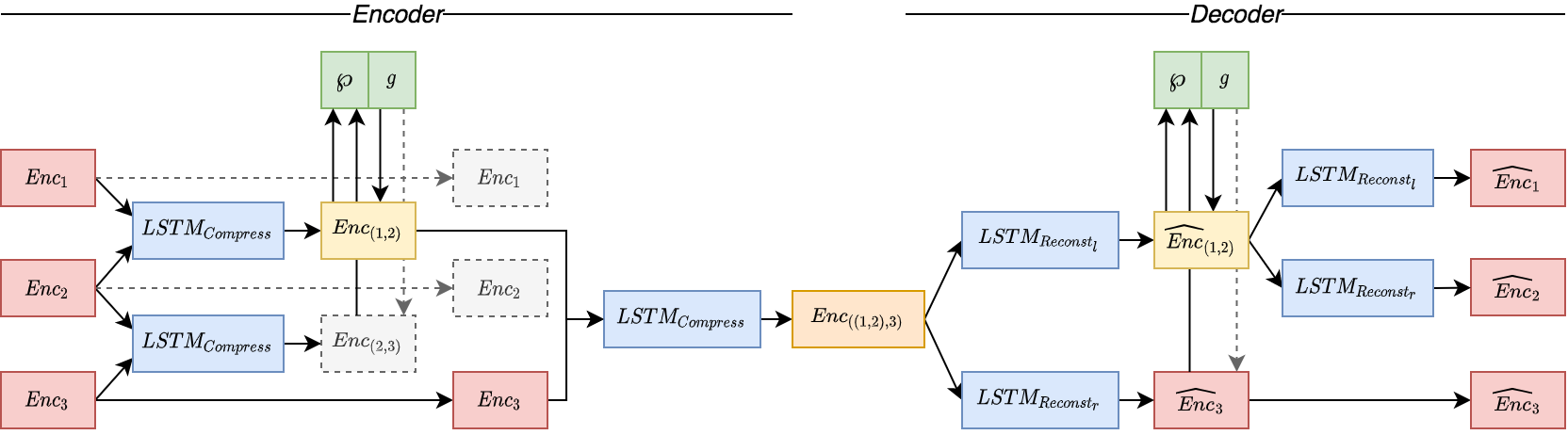}
    \caption{\textit{T-AE} (\textbf{T}ree-\textbf{A}uto\textbf{E}ncoder) topology for unsupervised tree inference. Inputs and outputs are dense encodings, $\widehat{Enc}_{x}$ represents reconstruction of spans. $\wp$ represents the pointer-network, $g \sim G(0,1)$ denotes the Gumbel-softmax (in the forward-pass with an additional straight-through computation, not shown here).\\ \colorbox{_grey}{Grey}/Dashed components represent actions outside the computational path chosen. \colorbox{_red}{red} = model inputs/outputs, \colorbox{_blue}{blue} = TreeLSTM cells, \colorbox{_green}{green} = discrete structure selector as in \citet{choi2018learning}, \colorbox{_yellow}{yellow} = hidden subtree encodings, \colorbox{_orange}{orange} = hidden state of the complete input.}
    \label{fig:topology}
\end{figure*}
The general task of tree inference has been mostly explored on sentence-level. For instance in \citet{choi2018learning} and \citet{socher2011semi} as described above, or by applying a reinforcement approach \cite{yogatama2016learning} or CKY methodology \cite{maillard2019jointly} to syntactic parsing. Our work employs a novel, fully differentiable approach to a similar problem in the area of discourse parsing. 

In discourse parsing itself, there have been multiple attempts to overcome the aforementioned limitations of small-scale human annotated datasets. However, all previous models (such as \citet{liu2018learning, huber2019predicting, liu2019single, huber2020mega}) use downstream tasks to infer discourse structures. While this is a valid strategy, shown to achieve SOTA results on the inter-domain discourse parsing task \cite{huber2020mega}, as well as performance gains on downstream tasks (e.g., \citet{liu2018learning, liu2019single}), those discourse structures are likely task-depended and need to be either combined across multiple downstream tasks 
or can only be applied in similar domains. 
Further work has been trying to infer RST-style discourse structures in a linguistically supervised manner 
\cite{nishida2020unsupervised}, showing 
good performance when heavily exploiting syntactic markers in combination with general linguistic priors. Yet, the approach appears to be very specific to the data at hand -- News articles from the Wall Street Journal -- raising questions in regards to overfitting.

In this work, we explore a purely unsupervised approach: instead of relying on domain specific syntactic features, we infer general discourse trees (structure only) 
by exploiting inherently available information from natural data (not requiring any supervision), making our model similar to approaches in language modelling \cite{jozefowicz2016exploring}. More specifically, our proposal extends the previously proposed Gumbel-TreeLSTM method \cite{choi2018learning} by substituting the original downstream-task related objective with an autoencoder-style reconstruction.

\section{Unsupervised Tree Autoencoder}
\label{model}
We now outline our general tree autoencoder model. The description is thereby purposely general, as the model is independent of a specific application and we believe can be utilized in manifold scenarios. 

Generally speaking, our proposed model induces tree structures through compression and reconstruction of raw inputs in a tree autoencoder style architecture. The model is similar in spirit to the commonly used sequence-to-sequence (seq2seq) architecture \cite{sutskever2014sequence}, which has also been interpreted as a sequential autoencoder \cite{li2015hierarchical}.
However, our approach generalizes on the seq2seq model, which is essentially a special (left-branching) case of a tree-structured autoencoder. While the sequential structure of a document 
is naturally given by the order of words, 
EDUs, and sentences
, moving towards more general tree representations adds the additional difficulty to infer valid tree structures alongside the hidden states.
To generate these discrete tree structures during training, in conjunction with the hidden states of the neural network, we make use of the Gumbel-softmax decision framework, allowing us to discretely generate tree-aggregations alongside intermediate sub-tree encodings \cite{Gumbel1948statistical, maddison2014sampling, jang2016categorical}. As presented in Figure \ref{fig:topology}, the structure of our novel \textit{T-AE} (\textbf{T}ree-\textbf{A}uto\textbf{E}ncoder) model comprises of 
an encoder, compressing the input into a fixed-size hidden vector and a subsequent decoder component, reconstructing the inputs in an autoencoder-style fashion. 

\subsection{Encoder Component}
The computational steps performed in our encoder are akin to the approach described in \citet{choi2018learning}, 
computing a single document encoding through a tree-style aggregation procedure. 
Our approach generates a hidden state $Enc_{l,r} = [c_p, h_p] = LSTM_{Compress}(l, r)$ for every two adjacent input embeddings $l = [c_l, h_l]$ (left) and $r = [c_r, h_r]$ (right) using a binary TreeLSTM cell as proposed by \citet{tai2015improved}\footnote{Equation \ref{eq:lstm} is modified from \citet{choi2018learning} and \citet{tai2015improved}.}. 

\begin{align}
\label{eq:lstm}
\begin{split}
    \begin{bmatrix} i \\ f_l \\ f_r \\ o \\ u \end{bmatrix} &=  \begin{bmatrix} \sigma \\ \sigma \\ \sigma \\ \sigma \\ tanh \end{bmatrix} \cdot (W \begin{bmatrix} h_l \\ h_r \end{bmatrix} + b)\\
    c_p &= f_l \cdot c_l + f_r \cdot c_r + i \cdot u\\
    h_p &= o \cdot tanh(c_p)
\end{split}
\end{align}

With $W \in  {\rm I\!R}^{5|h_p| \times 2|h_p|}$ and $b \in {\rm I\!R}^{2|h_p|}$. Based on the $(n-1)$ sub-tree candidates $Enc_{l,r}$ with $0 \leq l < (n-1)$ and $r = l+1$ of the given inputs $I$ ($|I| = n$), an un-normalized attention computation (or pointer network) $\wp = Pointer(\cdot,\cdot)$ \cite{vinyals2015pointer} is used to predict which two adjacent units should be merged. Randomly uniform Gumbel noise, obtained from the Gumbel distribution $G(0,1)$, effectively sampling $g \sim G(0,1)$ as $g_i=-log(-log(u_i))$ and $u_i=Uniform(0,1)$ is added to the un-normalized scores. Subsequently, the scores are normalized across aggregation candidates according to the temperature coefficient $\tau$ to obtain $p(l,r)$ (see equation \ref{eq:Gumbel}).

\begin{equation}
  \label{eq:Gumbel}
  p(l,r) = \ffrac{exp[(\wp(l,r)+g)/\tau]}{\sum_{k=0}^{n-1}{exp[(\wp(I_k,I_{k+1})+g)/\tau]}}
\end{equation}

In the forward pass, the straight-through (ST) Gumbel-distribution is used to enforce a discrete selection $p_{st}$, as commonly done using the Gumbel-softmax trick (see equation \ref{eq:st} \cite{jang2016categorical, choi2018learning, corro2018differentiable, corro2019learning}).

\begin{equation}
\label{eq:st}
    p_{st}(l,r) = 
    \begin{cases}
        1, & \text{if } \argmax{k=0,...,n-2}p(I_k, I_{k+1}) = l\\
        0, & \text{otherwise}
\end{cases}
\end{equation}

Given this one-hot encoding for a set of aggregation candidates, the most appropriate aggregation, as predicted by the pointer component and pertubed with the Gumbel-softmax, is executed. All other inputs with $p_{st}=0$ are directly forwarded to the next step and the respective TreeLSTM computations are discarded (grey/dashed boxes in Figure \ref{fig:topology}). In the example shown in Figure \ref{fig:topology}, Enc\textsubscript{1} and Enc\textsubscript{2} are aggregated, while Enc\textsubscript{3} is directly forwarded to the next step without any aggregation computation.

We recursively generate $n-1$ tree-candidates 
using the TreeLSTM cell in conjunction with the pointer-component and the Gumbel-softmax to build a discrete tree in bottom-up fashion, along with sub-tree hidden states\footnote{\label{footnote}Please note that the computation of the hidden states in the TreeLSTM cell and the tree structure prediction using the pointer-network with Gumbel pertubation are non-overlapping, allowing for independant optimization of either component.}. Once the tree is aggregated, a single hidden-state represents the complete input. 
Given this dense hidden-state (orange in Fig. \ref{fig:topology}), 
\citet{choi2018learning} add a multi-layer-perceptron (MLP) to predict the sentence-level sentiment on the Stanford Sentiment Treebank (SST) \cite{socher2013recursive}. As a result, the obtained tree structures are mostly task-dependent, as shown in \citet{williams2018latent}. With the goal to generate task-independent structures, we replace the task-dependant MLP layer with our autoencoder objective to reconstruct the original inputs. 

\subsection{Decoder Component}
The decoder component is implemented as an inverse TreeLSTM with two independent LSTM cells, recursively splitting hidden states into two separate encodings, reconstructing the left and right child-nodes (see equation \ref{eq:inverse_lstm}).

\begin{align}
\label{eq:inverse_lstm}
\begin{split}
    \begin{bmatrix} i_l \\ f_l \\ o_l \\ u_l \\ i_r \\ f_r \\ o_r \\ u_r \end{bmatrix} &=  \begin{bmatrix} \sigma \\ \sigma \\ \sigma \\ tanh \\ \sigma \\ \sigma \\ \sigma \\ tanh \end{bmatrix} \cdot (W h_p + b)\\
    c_l &= f_l \cdot c_p + i_l \cdot u_l\\
    c_r &= f_r \cdot c_p + i_r \cdot u_r\\
    h_l &= o_l \cdot tanh(c_l)\\
    h_r &= o_r \cdot tanh(c_r)
\end{split}
\end{align}

With $W \in  {\rm I\!R}^{8|h_p| \times |h_p|}$ and $b \in {\rm I\!R}^{|h_p|}$. Guided by the predicted tree structure of the ST Gumbel-softmax, as shown in Figure \ref{fig:topology} and equations \ref{eq:Gumbel} and \ref{eq:st}, the structural decision process in the reconstruction phase selects the highest scoring node to be further subdivided into a local sub-tree.
This reconstruction approach, generating two child-node encodings given the parent encoding $Enc_p \rightarrow [Enc_l, Enc_r]$ is recursively applied top-down until the original number of inputs $|I| = n$ is reached. Finally, the reconstructed dense encodings $[\widehat{Enc_1}, ..., \widehat{Enc_n}]$ are evaluated against the model input encodings, following the autoencoder objective.

\section{Discourse Tree Generation}
\label{discourse_adaptations}
The T-AE approach described above has been kept deliberately general. In this section, we 
outline the application-specific extensions required in order to deal with the inputs, assumptions, and granularity of the discourse parsing task.
First, for the task of discourse parsing, the model inputs $I$ are clause-like EDUs, representing sentence fragments containing multiple words. While the input- and output-encodings for word-level autoencoders are naturally represented as the respective one-hot vectors of words in the vocabulary, this approach is not directly applicable for discourse parsing
. Hence, we encode the EDUs as dense representations and execute the autoencoder objective directly on these embeddings \cite{press2016using}. 
Second, as discourse parsing considers complete documents, frequently containing a large number of sentences with oftentimes diverse content, we apply a commonly used approach in this area by separating within-sentence and between-sentence sub-trees \cite{joty2015codra}. In this setup, we apply the model described above for each sentence individually, trying to infer general patterns on sentence-level and subsequently using the learned sentence encodings (orange in Figure \ref{fig:topology}) as the starting point of the document-level T-AE. Having two separate models on sentence- and document-level further aligns with previous work in discourse parsing, postulating different sets of features relevant on different levels of the tree-generation process \cite{joty2015codra, wang2017two}.

\section{Evaluation}

\subsection{Tasks}
\label{tasks}
To fully evaluate the performance of our T-AE method, we conduct experiments on three distinct tasks, focusing on the two learning goals of our model:
\textbf{(1)} Evaluating if the model is able to infer valuable and general discourse-structures and
\textbf{(2)} Assessing the ability of the model to learn task-independent hidden states, capturing important relationships between instances. The three tasks are:\\
\textbf{Alignment with existing RST-style Discourse Structures:} Proposing an unsupervised approach to generate (discourse) tree structures allows us, in principle, to generate trees in any domain with sufficient raw text for training. However, due to the expensive and tedious annotation of gold-standard discourse trees, only very few datasets in some narrow domains are augmented with full RST-style trees, required to evaluate our generated structures. 
Despite their limited coverage, we believe that comparing the discourse-structures produced by our newly proposed model on the alignment with human-annotated discourse structures can be insightful.\\
\textbf{Ability to Predict Important Downstream Tasks:} Besides evaluating the overlap with existing, human-annotated discourse trees, we investigate into the generality of the T-AE model by evaluating the performance when applied to an important downstream task in NLP -- sentiment analysis. We therefore use the document-level hidden state of our model (orange in Figure \ref{fig:topology}), trained on the unsupervised autoencoder objective, and add a single feed-forward neural network layer on top, reducing the hidden state of our model to the number of sentiment classes required for the sentiment prediction task. Training this linear combination on top of the model's document-level encoding gives further insight into the information contained in the hidden state and its alignment with the downstream task of sentiment analysis.\\
\textbf{General Representational Consistency:} In this third task, we further explore the information captured by the document-level hidden state by qualitatively comparing the dense encoding of a random (short) sample document 
with its most similar/most different documents, giving intuition about the relatedness of similarly encoded documents.

\subsection{Datasets}
\textbf{The RST-DT Treebank} published by \citet{carlsonbuilding} is the most popular RST treebank. It contains 385 documents of the Wall Street Journal (WSJ) corpus, split into 344 documents in the training-set and 39 documents in the test-portion. In order to obtain a development set, we subdivide the training-portion into 308 documents for training and 36 documents for a length-stratified development set. Each document in the RST-DT treebank is annotated with a complete discourse tree according to the RST discourse theory and segmented into EDUs by human annotators. N-ary subtrees are converted into a sequence of right-branching constituents.\\
\textbf{The Yelp'13 Dataset} by \citet{tang2015document} is a review dataset published as part of the 2013 Yelp Dataset Challenge. The corpus contains predominantly restaurant reviews alongside a 5-point star rating. Frequently used in previous work, the dataset has been pre-segmented into EDUs by \citet{angelidis2018multiple}, using the discourse-segmenter proposed in \citet{feng2012text}. The complete dataset contains 335,018 documents in an 80-10-10 data-split, resulting in 268,014 training documents and 33,502 documents each in the development and test sets.
\\


\subsection{Baselines}
\label{baselines}
For the \textbf{Alignment with RST-style Discourse Structures}, we evaluate three sets of related approaches: For supervised models, we compare against a diverse set of previously proposed, fully supervised discourse parsers, trained and evaluated on the RST-DT dataset. 
These include the CODRA model by \citet{joty2015codra}, the Two-Stage approach by \citet{wang2017two} and the neural topology by \citet{guz-etal-2020-unleashing}.
We further compare 
against our two distantly supervised models recently proposed in \citet{huber2019predicting, huber2020mega}, using sentiment analysis to inform the generation of discourse structures with distant supervision. 
Our last set of baselines for this task contains linguistically supervised approaches. We compare our model against fully left- and right-branching trees, as well as hierarchically left- and right-branching tree structures (separated on sentence-level), encoding basic rhetorical strategies. 
Left-branching trees generally reflect a common sequential strategy while 
right-branching tree structures oftentimes accurately represent documents where the main objective is initially expressed and then further evaluated throughout the document (e.g. news)\footnote{Right-branching trees are further artificially favoured in discourse parsing, since most parsing models convert n-ary sub-trees into a sequence of right-branching constituents.}.
For these reasons, we consider the left- and right-branching tree structures as linguistically supervised approaches. 
We further show the recently proposed model by \citet{nishida2020unsupervised} in our evaluation. In their best model setting, \citet{nishida2020unsupervised} also heavily exploit basic rhetorical strategies of natural language by aggregating a document into right-branching trees on sentence- and paragraph-level and joining paragraphs using left-branching constituents. Starting from this linguistically inspired tree (already achieving remarkable performance on the well-structured news documents), they apply a Viterbi EM algorithm to achieve further improvements. Despite the promising results on RST-DT, we believe that such high performance is mostly due to the well-structured nature of news documents and not generally applicable to other domains -- the main objective of our presented approach.

Building on the intuition given in \citet{huber2019predicting, huber-carenini-2020-sentiment,  huber2020mega}, we further evaluate our model regarding the \textbf{Ability to Predict Important Downstream Tasks}. More precisely, we evaluate the sentiment prediction performance of the document-level hidden-state of our model
against 
the HAN model proposed by \citet{yang2016hierarchical}, the LSTM-GRNN approach by \citet{tang2015document} as well as a document encoding build from average random word encodings and the majority class baseline. 


\begin{table}[t!]
\centering
\setlength{\belowcaptionskip}{-10pt}
\scalebox{1}{
{\renewcommand{\arraystretch}{1}
\begin{tabular}{|l|r | r|}
\hline
Model & Structure\\
\hline \hline 
\textbf{Human} \shortcite{morey2017much} & 88.30 \\
\hline\hline
\multicolumn{2}{|c|}{\textbf{Supervised}}\\
\hline
CODRA\shortcite{joty2015codra} & 83.84 \\
Two-Stage\shortcite{wang2017two} & 86.00 \\
Neural-SR\shortcite{guz-etal-2020-unleashing} & \textbf{86.47} \\
\hline \hline
\multicolumn{2}{|c|}{\textbf{Distantly Supervised}}\\
\hline
Two-Stage\textsubscript{Yelp13-DT}\shortcite{huber2019predicting} & 76.41 \\
Two-Stage\textsubscript{MEGA-DT}\shortcite{huber2020mega} & \textbf{77.82}\\
\hline \hline
\multicolumn{2}{|c|}{\textbf{Linguistically Supervised}}\\
\hline
Left Branching & 53.73 \\
Right Branching & 54.64 \\
Hier. Left Branching & 70.58 \\
Hier. Right Branching & 74.37 \\
ViterbiEM\shortcite{nishida2020unsupervised} & \textbf{84.30} \\
\hline \hline
\multicolumn{2}{|c|}{\textbf{Unsupervised}}\\
\hline
Ours\textsubscript{RST-DT} & 69.68 \\
Ours\textsubscript{Yelp'13} & \textbf{71.32} \\
\hline
\end{tabular}}}
\caption{Results of the average micro-precision measure, evaluated on the RST-DT corpus. Subscripts identify training sets. Best model in each subset is \textbf{bold}.}
\label{tab:rst}
\end{table}




\subsection{Hyper-Parameters Settings}
We select our hyper-parameters based on the development-set performance of the respective datasets. Despite the fact that we are training two unsupervised models (on RST-DT and Yelp'13) we use a single set of hyper-parameters, to be more general. We train all models using the Adam optimizer \cite{kingma2014adam} with the standard learning rate of $0.001$. As mentioned before, we are directly training on dense representations of input EDU embeddings, comparing them to the reconstructed representations of EDUs. This setup makes the Kullback-Leibler Divergence (KLD) or the Mean-Squared-Error (MSE) the natural choice for the loss function. In this work we employ MSE due to its superior performance observed on the development-set. Each EDU in the input document is represented as the average GloVe word-embedding \cite{pennington2014glove} as in \citet{choi2018learning}. The loss is computed on the softmax of the respective inputs and outputs. We train our model on mini-batches of size $20$, due to computational restrictions\footnote{Trained on a Nvidia GTX 1080 Ti GPU with 11GB of memory.} and apply regularization in form of $20\%$ dropout on the input embeddings, the document-level hidden state and the output embeddings \cite{choi2018learning}. We clip gradients to a max norm of $2.0$ to avoid exploding gradients. Documents are limited to $150$ EDUs per document and a maximum of 50 words per EDU, similar to \citet{huber2019predicting}. We restrict the vocabulary size to the most frequent $50,000$ words with an additional minimal frequency requirement of $10$. We train the sentence- and document-level model for $40$ epochs and select the best performing generation on the development set. The hidden dimension of our LSTM modules as well as the pointer component is set to $64$, due to computational restrictions. To avoid our model to interfere with the input GloVe embeddings, we freeze the word representations. To promote consistency between the encoding and decoding, we tie the decoder tree-decisions to the encoder predictions, enabling a more consistent tree-embedding in the compression and reconstruction phase. Furthermore, to disentangle the optimization of structures and hidden states, we apply a phased approach, alternating the training of the two components in a conditional back-propagation loop with a single objective in each pass over the data (see footnote \ref{footnote}). This way, the hidden states are recalculated based on the last epoch's structure prediction and vice-versa. To be able to explore diverse tree candidates in early epochs and further improve them during later epochs, we start with the diversity factor $\tau = 5$ and linearly reduce the parameter to $\tau = 1$ (see \citet{choi2018learning}) over $3$ structure-learning epochs.

\begin{table}[t!]
\centering
{\renewcommand{\arraystretch}{1.1}
\setlength{\belowcaptionskip}{-10pt}
\scalebox{1}{
\begin{tabular}{|l|r|}
\hline
Model & Accuracy\\
\hline \hline 
HAN\shortcite{yang2016hierarchical} & \textbf{66.20} \\
LSTM-GRNN\shortcite{tang2015document} & 65.10\\
Ours\textsubscript{Yelp'13} & 42.69 \\
Ours\textsubscript{RST-DT} & 40.41 \\
Random Encoding & 37.30\\
Majority Class & 35.63\\
\hline
\end{tabular}}}
\caption{Five-class sentiment accuracy scores trained and tested on the Yelp'13 dataset, subscripts in model-names indicate dataset for unsupervised training. Best model is \textbf{bold}.}
\label{tab:sentiment}
\end{table}

\begin{table*}[ht!]
    \centering
    {\renewcommand{\arraystretch}{1.1}
    \setlength{\belowcaptionskip}{-10pt}
    \scalebox{.98}{
    \begin{tabular}{| c | p{15cm} |}
        \hline
        \textbf{Document} & Prices were cheap, however food was served well after others who came in and they literally put brown gravy on the Mexican food, staff ignored simple requests. Only reason for 1 star was due to price. \\
        \hline\hline
        \textbf{Similar-1} & This establishment has a good 10\$ lunch special with plenty of varity in the bento they offer, service is usually good polite and efficient the only thing that makes me crazy is the crappy usually too loud caned pop music they play. \\
         \hline
        \textbf{Similar-2} & The good: Awesome complimentary breakfasts, warm gooey chocolate chip cookie at check in, nice pool and and hot tub at the center, fairly large room with 2 TVs (flat screen) and a huge comfortable bed with down pillows. The bad: Not a bad fitness room but could be larger, it doesn't have the feel look of a fancy hotel at first. More like a Motel (but the rooms are nice and the restaurant too). The ugly: No free internet \\
         \hline
        \textbf{Similar-3} & Just the facts: Great options for healthier eating, unique non-meat sandwich options at lunch (portabello, grilled zucchini, black bean, etc.), decent coffee, cute atmosphere and fun s\&p shakers at the table, kind of pricey. I want to go back to try breakfast. \\
         \hline\hline
        \textbf{Different-1} & Forgot to mention the prices are great \& just had the baklava yummm to die for delicious! \\
        \hline
        \textbf{Different-2} & Bit pricey, but it's always been our favorite place to go for treats. \\
        \hline
        \textbf{Different-3} & Decent place, but the drinks are too expensive unless its a buy 1 get 1 night. \\
         \hline
    \end{tabular}}}
    \caption{Representationally similar/different document-encodings based on the cosine similarity. For more examples of the representational similarity and additional tree structure comparisons see Appendix \ref{representations} and \ref{trees} respectively.}
    \label{tab:semantic_similarity}
\end{table*}

\subsection{Experiments}
In this section we evaluate our novel T-AE model on the three tasks described in section \ref{tasks}. 
Table \ref{tab:rst} shows the results on the first task, evaluating our model on RST-style discourse structures from the RST-DT treebank.
The first sub-table shows three top-performing, completely supervised models, reaching a structure-prediction performance of $86.47\%$ using the neural approach by \citet{guz-etal-2020-unleashing}. 
In comparison, the second sub-table contains our distantly supervised models, achieving a performance of $77.82\%$ \cite{huber2020mega}. 
The third sub-table presents the linguistically supervised models, showing a clear advantage of the right-branching models over left-branching approaches, in line with our intuition given in section \ref{baselines}. Furthermore, considering sentence boundaries and generating hierarchical baselines significantly improves the performance, reaching $74.37\%$ with the hierarchical right branching baseline and $70.58\%$ on the left-branching structures. The linguistically supervised Viterbi EM approach by \citet{nishida2020unsupervised} reaches a performance of $84.30\%$ with their multi-level hierarchical approach.
Our newly proposed, truly unsupervised and purely data-driven approach is shown in the fourth sub-table. In comparison to the aforementioned linguistically supervised models, this set of results makes no assumptions on the underlying data except the sentence/document split. When trained on the raw-text of the small-scale RST-DT dataset, our T-AE approach reaches a performance of $69.68\%$, slightly below the linguistically supervised hierarchical left-branching model. Even though the unsupervised training corpus is within the same domain as the test dataset, the very limited amount of data seems insufficient for the unsupervised model. Training our model on the nearly three orders of magnitude larger Yelp'13 dataset, we reach a performance of $71.32\%$ evaluating the tree structures on RST-DT. This result shows that a larger training dataset, even though containing out-of-domain documents (reviews vs. news), can improve the performance over the within-domain model trained on a small-scale dataset and the hierarchical left-branching model. 

To evaluate the ability of our model to capture valid information to represent input documents, we assess the document-level hidden state's ability to capture useful information for the downstream task of sentiment analysis. The results of this experiment are provided in Table \ref{tab:sentiment}, showing the accuracy of our models when compared against commonly used approaches. The best system (the HAN model) reaches an accuracy of $66.2\%$, while the random baseline reaches $37.30\%$ and the simple majority class baseline achieves $35.63\%$. Our models based on the T-AE hidden states obtain accuracy scores in-between those results, reaching $40.41\%$ and $42.69\%$ when trained on RST-DT and the much larger Yelp'13 respectively. While this performance is still far from the results of completely supervised models, the improvements over the simple baselines suggest the usefulness of our learned document-level encodings.

In our third and last experiment, we aim to further evaluate the quality of the document encodings in a qualitative manner. We therefore compare the hidden-state of a random document from the Yelp'13 test-set against all datapoints in the test-portion and show the three most similar/most different documents according to the cosine similarity measure in Table \ref{tab:semantic_similarity}. It can be observed that closely related documents have a similar argumentative structure as the core-document (top row in Table \ref{tab:semantic_similarity}), initially describing a positive aspect and subsequently evaluating on negative components. The most different documents tend to have an inverse structure.


%
%
%


\section{Conclusion and Future Work}
In this paper, we proposed a truly unsupervised and purely data-driven tree-style autoencoder to compress and reconstruct textual data. We show the potential of our T-AE approach on the task of discourse parsing, which severely suffers from training-data sparsity, due to the tedious and expensive annotation process. Our unsupervised model outperforms one of the commonly used, linguistically supervised approaches, without making any assumptions on the underlying data, except the sentence/document split. The superior performance compared to the hierarchical left branching baseline plausibly indicates that our unsupervised structures could be valuable when combined with supervised or distantly supervised models to further improve their joint performance. Furthermore, the superior performance of the large out-of-domain model trained on the Yelp'13 dataset over the small-scale within-domain model trained on the raw text of the RST-DT dataset shows the synergies between these corpora as well as strong potential for even larger datasets to enhance the performance of the approach. 

In the future, we intend to extend this work in several ways: First, we want to explore the application of generative models, employing a variational autoencoder. Second, we plan to study further tasks besides predicting discourse, such as syntactic parsing, as well as additional synergistic downstream tasks (e.g. summarization, text classification). To improve our model on important downstream tasks (such as sentiment analysis), we want to explore a pre-training/fine-tuning approach, similar to contextualized language  models, such as BERT. Combining our novel approach with distantly-supervised and supervised models is another future direction we want to explore. Lastly, we plan to evaluate additional model adaptions, such as two independent models on sentence- and document-level, incorporating a BERT EDU encoder and an end-to-end model with soft-constraints on sentence-level.

\section*{Acknowledgments}
We thank the anonymous reviewers, Linzi Xing and the UBC-NLP group for their comments and suggestions. \\
This research was supported by the Language \& Speech Innovation Lab of Cloud BU, Huawei Technologies Co., Ltd. \\
We further acknowledge the support of the Natural Sciences and Engineering Research Council of Canada (NSERC).\\
Nous remercions le Conseil de recherches en sciences naturelles et en génie du Canada (CRSNG) de son soutien.

\bibliography{aaai21}

\newpage

\appendix

\onecolumn

\section{Representational Consistency}
\label{representations}
We show three more random (short) examples of similar/different document encodings retrieved from our system using the Yelp'13 dataset for training. Example core-documents are limited to reviews with less than 50 words for readability.

\begin{table*}[h!]
    \centering
    {\renewcommand{\arraystretch}{1.1}
    \scalebox{.9}{
    \begin{tabular}{| c | p{15cm} |}
        \hline
        \textbf{Document} & Horrible service, they keep delaying the delivery of our order, they say they arrived and called with no answer to buy more time, when called to complain got hang up on, do not order from here. PS: I'm not an ex-employee.\\
        \hline\hline
        \textbf{Similar-1} & I used to love this place but it's cash only now. So inconvenient that if I want bahama bucks I drive further to go to one that accepts credit/debit.\\
         \hline
        \textbf{Similar-2} & Vito's is pretty good food, not the top on my list but my husband and I will go here for a special occasion from time to time when we cant get in elsewhere and it does the job.\\
         \hline
        \textbf{Similar-3} & Love this place! Great healthy choices and awesome desert. Way to go in healthy choices!\\
         \hline\hline
        \textbf{Different-1} &  Love the Happy Hour at Brio's . Great food, drinks and people to meet. The staff is very attentive and want to ensure you enjoy your time there.\\
        \hline
        \textbf{Different-2} & I love the Gyro's pita and I like the service, so I will be back to try some other things in the Menu. Good place!\\
        \hline
        \textbf{Different-3} & 
        Great menu of food, you really can't go wrong with any choice. Love how the chef can bring the spice. Definitely one of the better Asian restaurants in Scottsdale.\\
         \hline
    \end{tabular}}}
    \caption{Representationally similar/different document-encodings based on the cosine similarity.}
\end{table*}

\begin{table*}[h!]
    \centering
    {\renewcommand{\arraystretch}{1.1}
    \scalebox{.9}{
    \begin{tabular}{| c | p{15cm} |}
        \hline
        \textbf{Document} & 5 major health code violations. Its a shame because I really liked this place and the food was pretty good, but i wont be eating there again.\\
        \hline\hline
        \textbf{Similar-1} & Food is alright. Service is garbage. You would expect a lot more from a place with the kind of prices they have. If anything, the food is what gave it the 2 stars.\\
         \hline
        \textbf{Similar-2} & Had a not-so-great experience at US-egg. My veggie frittata was really soggy, like a veggie frittata soup. The frozen vegetables must have thawed in the dish.\\
         \hline
        \textbf{Similar-3} & Very cool, great music. Study room but beware of the smoke cloud surrounding the outside. Can't sit outside unless you want to leave smelling like smoke.\\
         \hline\hline
        \textbf{Different-1} & First time trying out this place and good thing I did. Good, helpful service and a quick turn around. There was a 50 cent debit charge but was waived off for being a first timer. Pretty awesome, not that many businesses would do that.\\
        \hline
        \textbf{Different-2} & Overall, it is a department store. You have helpful employees usually however when its comes to sale time, be ready to get in and get out. The customer to employee ratio is not the best and to achieve any assistance is nearly impossible.\\
        \hline
        \textbf{Different-3} & Beef ribs were perfectly spicy and sweet. The Indian fried bread and biscuits with whipped cinnamon butter are great. I loved the whipped butter served room temperature.\\
         \hline
    \end{tabular}}}
    \caption{Representationally similar/different document-encodings based on the cosine similarity.}
\end{table*}

\begin{table*}[h!]
    \centering
    {\renewcommand{\arraystretch}{1.1}
    \scalebox{.9}{
    \begin{tabular}{| c | p{15cm} |}
        \hline
        \textbf{Document} & Ask them to wear a hair net when they make your sandwich. I used to eat here a lot till I got hair in 3 sandwich's. Two where on the same day. Hell no will I go back.\\
        \hline\hline
        \textbf{Similar-1} & Worst experience ever! Salsa was spoiled and my chicken taco was disgusting. Even the little I did eat I got food poisoning. Will never return to this restaurant.\\
         \hline
        \textbf{Similar-2} & Service was marginal and food was not very good. I remember this place being better, but I guess they changed their Menu around a bit ago. Let's just say I like the old one better. I will pass on this place moving forward.\\
         \hline
        \textbf{Similar-3} & It's a fast food joint, that being said everything here went off without a hitch. They even brought our order to us at the table. That's something I've never seen done by this chain. I hope they keep up the good work.\\
         \hline\hline
        \textbf{Different-1} & Finally a chop shop in Tempe! The set up here is adorable and they have great healthy food. They've added a few new juices to the menu which I'm excited to try. Will definitely be here all the time!\\
        \hline
        \textbf{Different-2} & Perfect early Thai dinner. Spicy mussels. Tad pad with shrimp and chicken, perfect fit for a early dinner. No rush crowd here if you arrive a week night early dinner good food, great price!\\
        \hline
        \textbf{Different-3} & Good wholesome BBQ, but they were out of collard greens: the cornbread muffins weren't the best either, but the brisket and cowboy beans are worth coming for.\\
         \hline
    \end{tabular}}}
    \caption{Representationally similar/different document-encodings based on the cosine similarity.}
\end{table*}

\section{Tree Comparison}
\label{trees}
We compare our generated discourse trees (left) using the T-AE framework against gold-standard RST discourse trees (right) annotated by linguistic experts. We find that while there are differences between the tree structures, in the majority of cases the general tree structures overlap significantly.

\begin{figure*}[h!]
    \centering
    \begin{subfigure}[b]{0.45\textwidth}
         \centering
         \includegraphics[width=\textwidth]{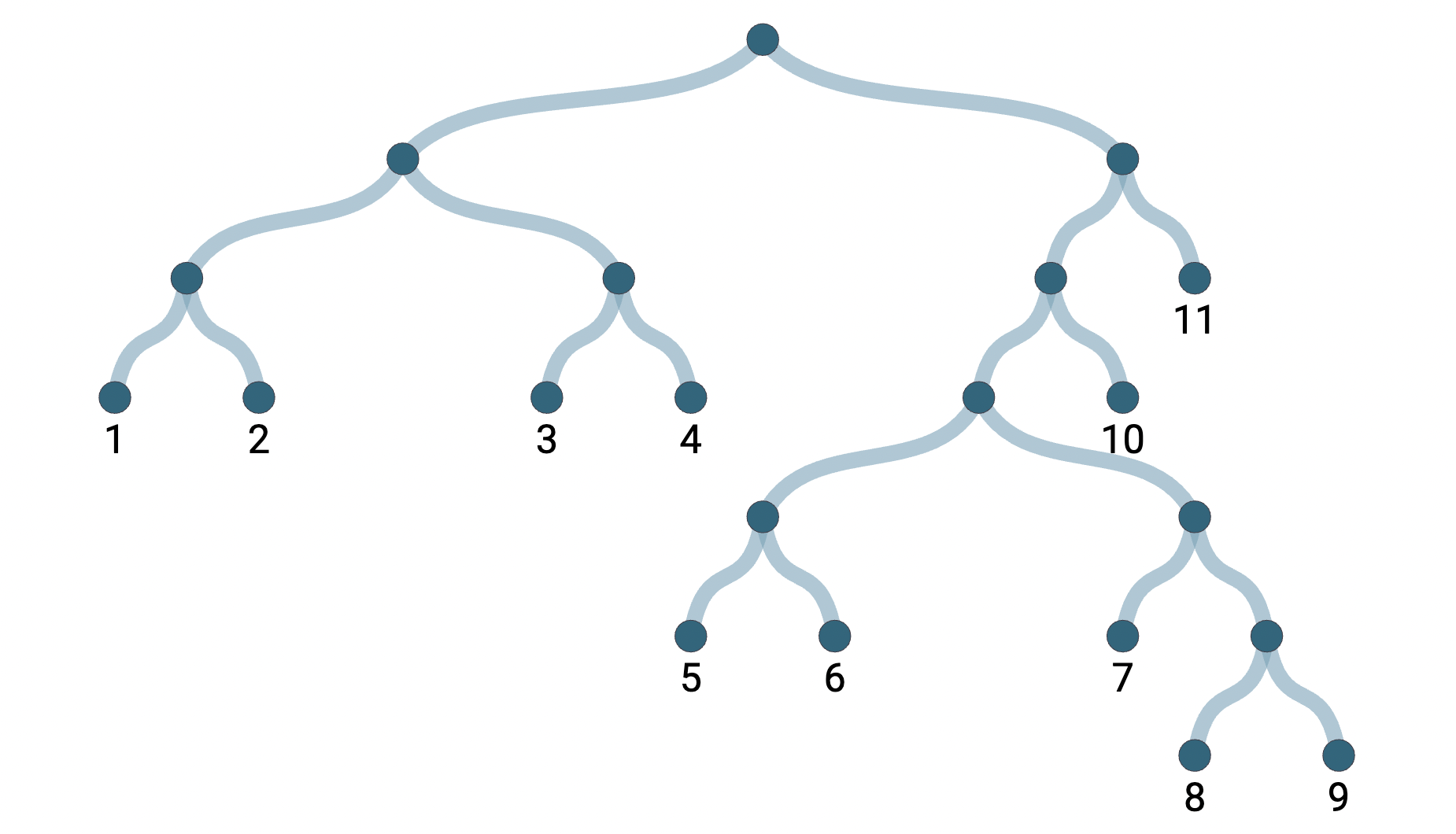}
     \end{subfigure}
     \rulesep
     \begin{subfigure}[b]{0.45\textwidth}
         \centering
         \includegraphics[width=\textwidth]{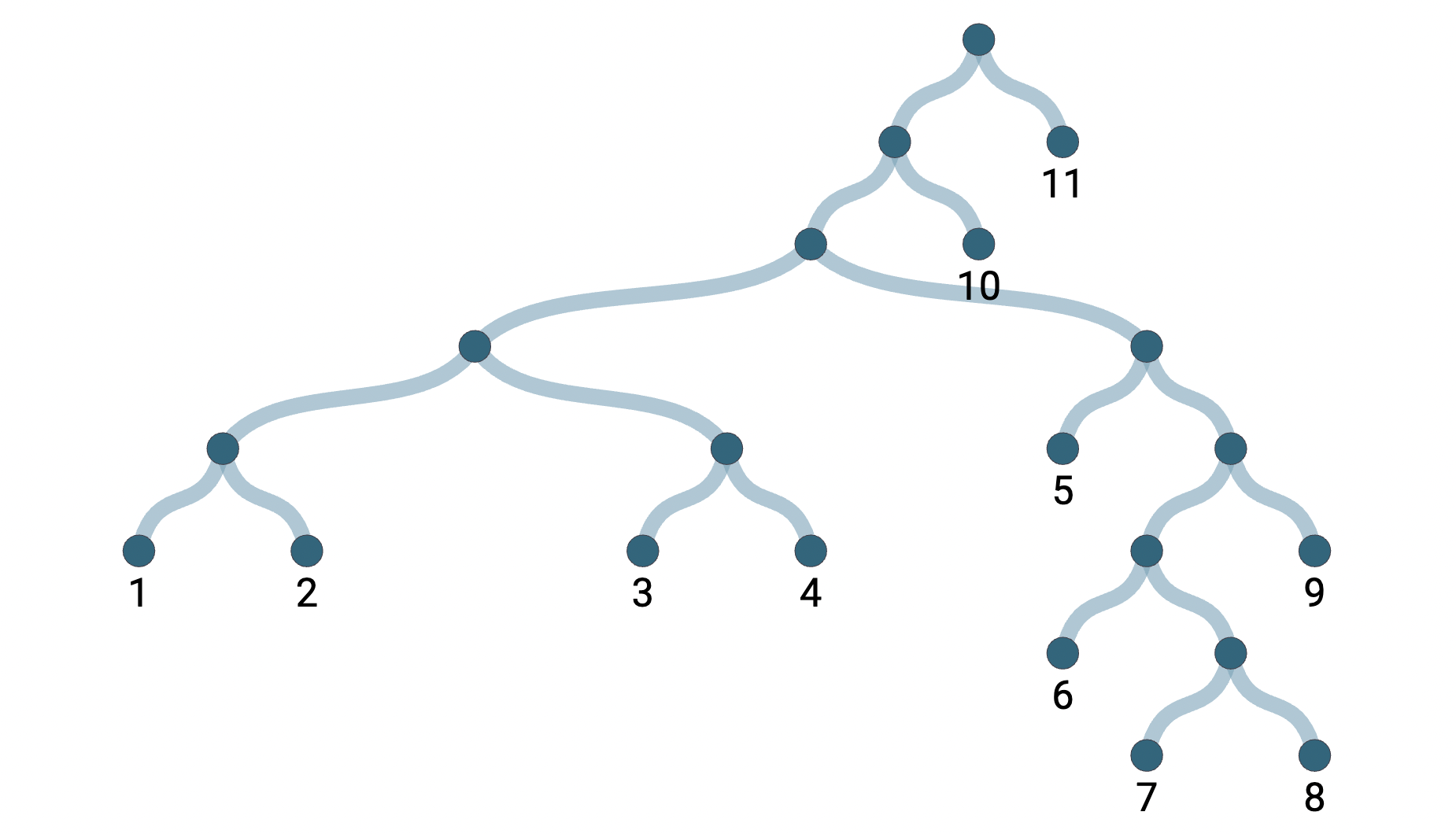}
     \end{subfigure}
    \caption{Our generated tree (left) compared to the gold-standard tree (right) for document \textit{wsj\_1395}}
\end{figure*}

\begin{figure*}[h!]
    \centering
    \begin{subfigure}[b]{0.45\textwidth}
         \centering
         \includegraphics[width=\textwidth]{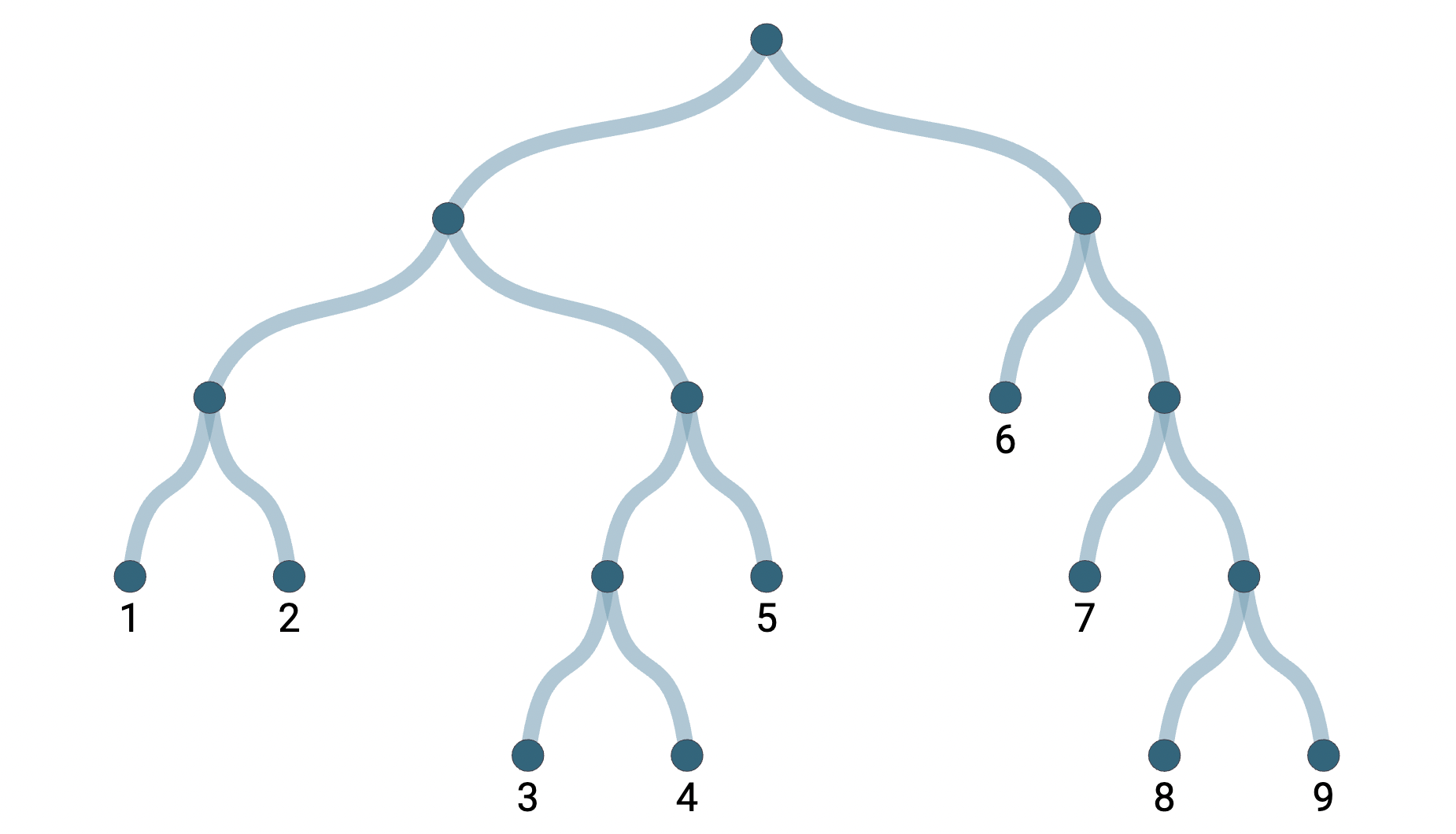}
     \end{subfigure}
     \rulesep
     \begin{subfigure}[b]{0.45\textwidth}
         \centering
         \includegraphics[width=\textwidth]{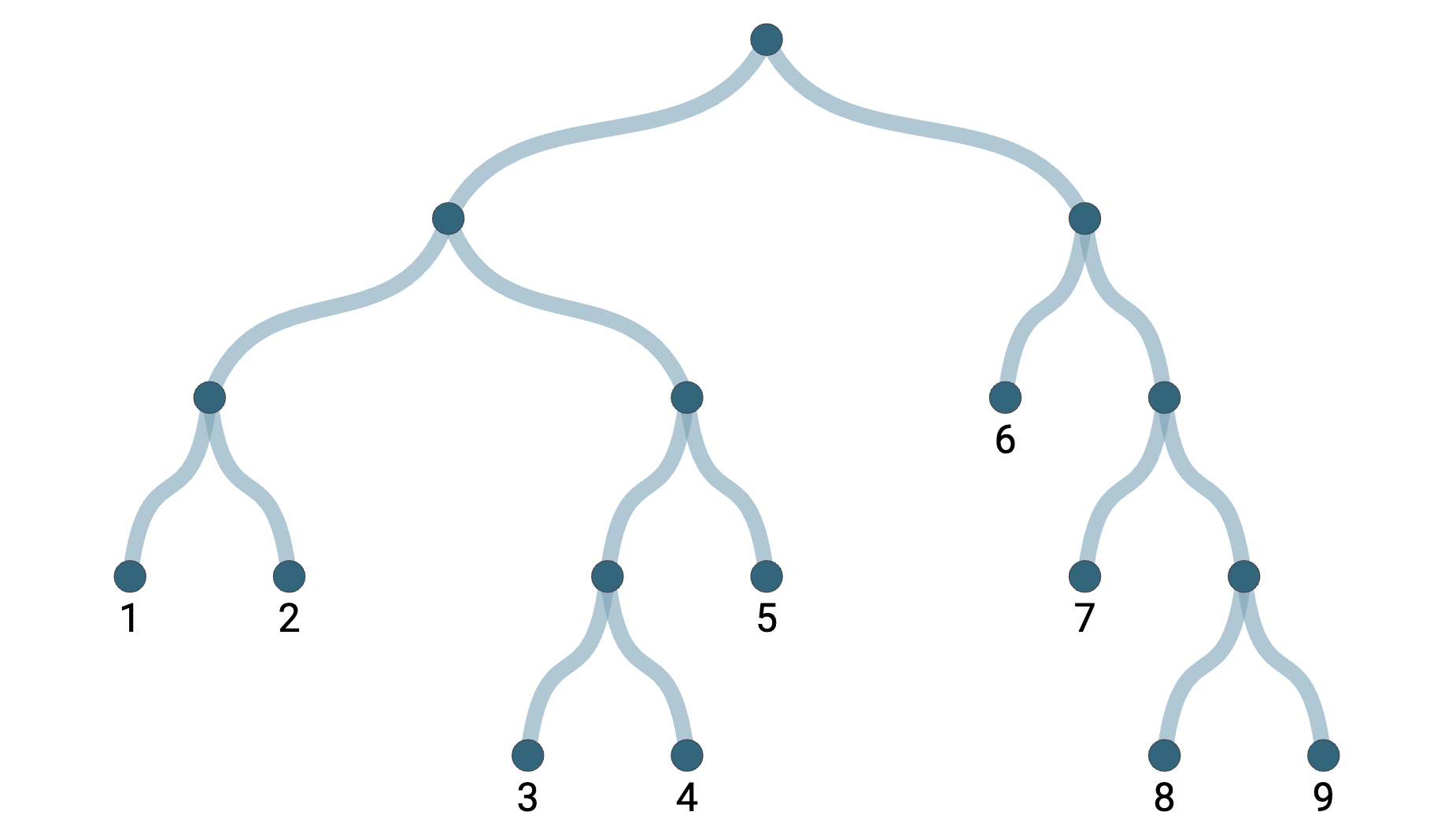}
     \end{subfigure}
    \caption{Our generated tree (left) compared to the gold-standard tree (right) for document \textit{wsj\_1198}}
\end{figure*}

\begin{figure*}[h!]
    \centering
    \begin{subfigure}[b]{0.45\textwidth}
         \centering
         \includegraphics[width=\textwidth]{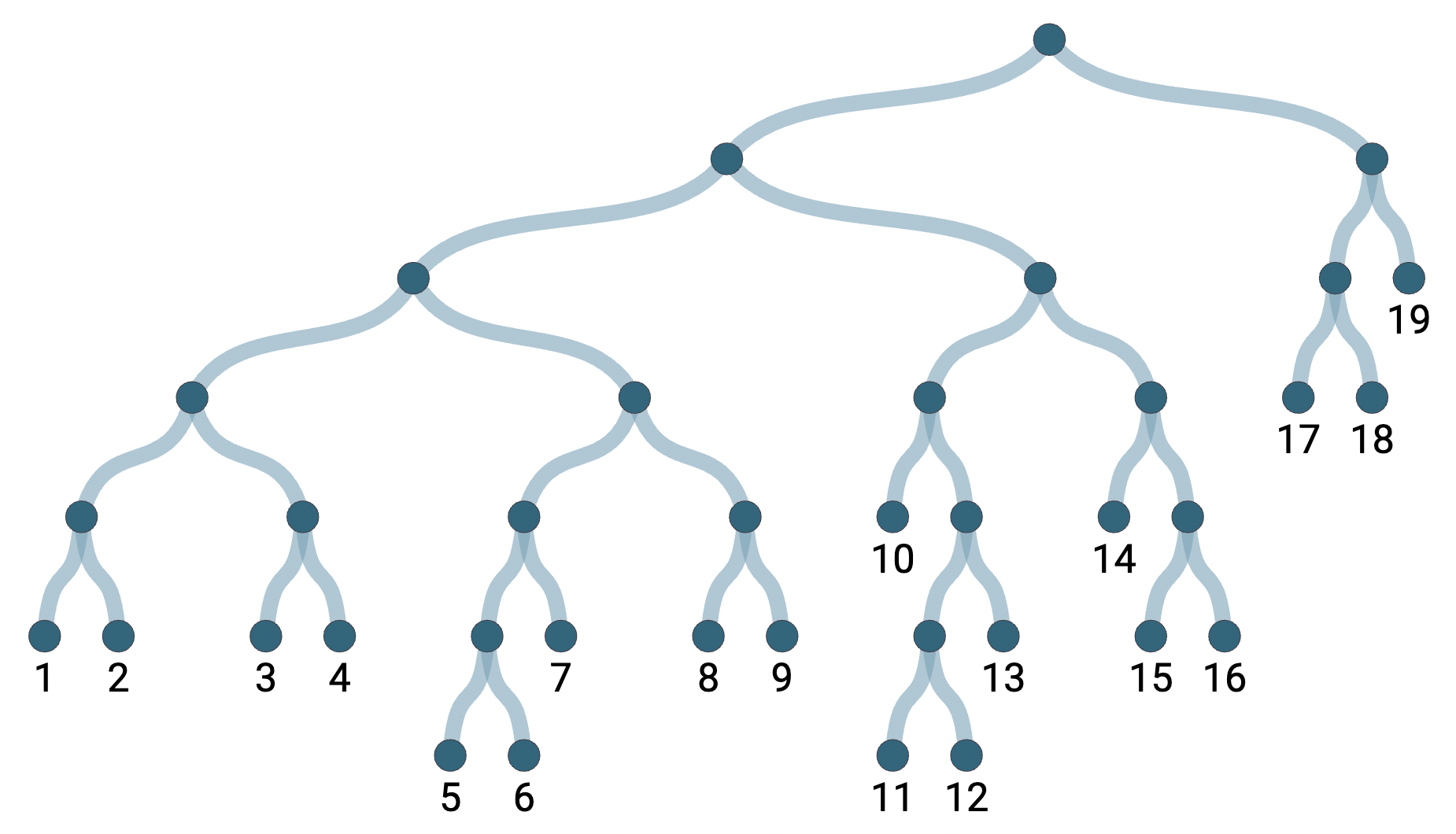}
     \end{subfigure}
     \rulesep
     \begin{subfigure}[b]{0.45\textwidth}
         \centering
         \includegraphics[width=\textwidth]{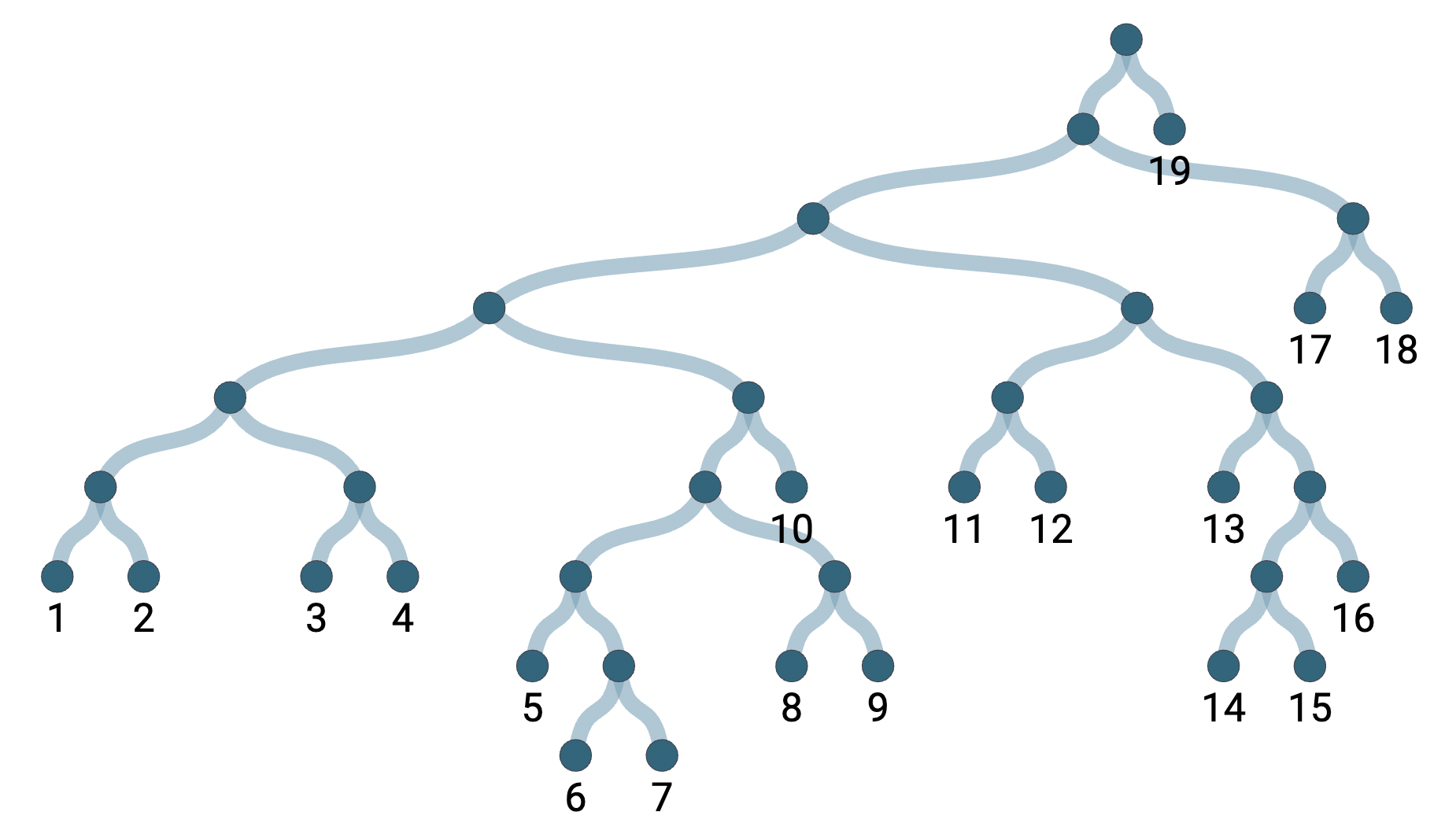}
     \end{subfigure}
    \caption{Our generated tree (left) compared to the gold-standard tree (right) for document \textit{wsj\_1998}}
\end{figure*}


\end{document}